\documentclass[runningheads]{llncs}

\usepackage{eccv}

\usepackage{eccvabbrv}

\usepackage{graphicx}
\usepackage{booktabs}
\usepackage{pgfplots}
\usepackage[accsupp]{axessibility}  

\usepackage{hyperref}

\usepackage{orcidlink}
\usepackage{multirow}

\newcommand\Ds{{\cal D}_S}
\newcommand\Du{{\cal D}_U}
\DeclareMathOperator*\argmax{\arg\,\max}

\begin{document}

\title{Are CLIP features all you need for Universal Synthetic Image Origin Attribution?} 

\titlerunning{Universal Synthetic Image Origin Attribution}

\author{Dario Cioni\inst{1,3}\orcidlink{0000-0003-4475-0981} \and
Christos Tzelepis\inst{2}\orcidlink{0000-0002-2036-9089} \and
Lorenzo Seidenari\inst{1}\orcidlink{0000-0003-4816-0268} \and
Ioannis Patras\inst{3}\orcidlink{0000-0003-3913-4738}}

\authorrunning{D.~Cioni~\etal}

\institute{University of Florence \and
City, University of London \and 
Queen Mary, University of London}

\maketitle

\begin{abstract}
The steady improvement of Diffusion Models for visual synthesis has given rise to many new and interesting use cases of synthetic images but also has raised concerns about their potential abuse, which poses significant societal threats. To address this, fake images need to be detected and attributed to their source model, and given the frequent release of new generators, realistic applications need to consider an Open-Set scenario where some models are unseen at training time. Existing forensic techniques are either limited to Closed-Set settings or to GAN-generated images, relying on fragile frequency-based ``fingerprint'' features. By contrast, we propose a simple yet effective framework that incorporates features from large pre-trained foundation models to perform Open-Set origin attribution of synthetic images produced by various generative models, including Diffusion Models. We show that our method leads to remarkable attribution performance, even in the low-data regime, exceeding the performance of existing methods and generalizes better on images obtained from a diverse set of architectures. We make the code publicly available at: \url{https://github.com/ciodar/UniversalAttribution}.
\end{abstract}

\section{Introduction}\label{sec:intro}
    
    Recent generative models can generate synthetic media of exceptional quality and diversity~\cite{esser2024scaling, imagen2, midjourney2022}, allowing the use of AI-Generated Content (AIGC) in a wide range of areas, such as medical image analysis~\cite{songsolving, ozbey2023unsupervised, marimont2023disyre, naval2024ensembled}, face reenactment~\cite{bounareli2023hyperreenact, bounareli2023stylemask, bounareli2024one, bounareli2024diffusionact}, image editing~\cite{tzelepis2021warpedganspace, oldfield2023panda, oldfield2024bilinear, tzelepis2022contraclip, oldfield2023parts, d2024improving}, face anonymization~\cite{barattin2023attribute}, art~\cite{cioni2023diffusion}, and fashion~\cite{morelli2023ladi}. However, alongside their benefits, the broad adoption of AIGC tools raises critical concerns about negative social impacts, including the exploitation of AIGC under malicious intent or towards intellectual property (IP) infringement. That is, users equipped with such powerful and readily available generative models can generate images with biased or inappropriate content and distribute them across digital platforms. Moreover, the unauthorized exploitation of generative models -- where model parameters are illicitly obtained and utilized for unintended commercial purposes -- poses a significant challenge to IP protection. Addressing the above problems requires the ability to i) detect synthetic content and ii) trace its provenance to the specific generative model responsible for its creation -- this task is typically referred to as \textit{attribution}~\cite{yang_deepfake_2022, bui_repmix_2022, yang_progressive_2023}.
    
    Whilst recent generative models span a wide range of architectures, they can be categorized into the following three major classes: i) Variational Autoencoders (VAEs)~\cite{ramesh2021zero}, ii) Generative Adversarial Networks (GANs)~\cite{brock2018large}, and iii) Diffusion Models (DMs)~\cite{esser2024scaling}. The attribution problem has mainly been studied in the case of GANs, where GAN-generated images are attempted to be attributed to the specific GAN architecture that gave rise to them. For doing so, early works (e.g.,~\cite{marra_gans_2019}) relied on distinctive residual features, typically referred to as \textit{fingerprints}, that are inherited in the image during the generation process, and which can be extracted by performing a frequency transformation of the image~\cite{marra_gans_2019, yu_attributing_2019}. Further research in GAN attribution~\cite{zhang_detecting_2019, frank2020leveraging, durall_watch_2020} validated the existence of such fingerprints, allowing to identify the generator among a finite, fixed (i.e., closed) set of models. Albeit, the fast progress and the rapid introduction of new generative models (both under the GAN- and diffusion-based paradigms) greatly limits the applicability of such closed-set setting. A few recent works in GAN attribution addressed this limitation by allowing the attribution at an architecture-level~\cite{yang_deepfake_2022, bui_repmix_2022} or by adopting an Open Set Recognition approach~\cite{yang_progressive_2023}. However, such methods do not take into account diffusion-based generative models, which have emerged as the leading generative paradigm, whilst at the same time the adopted fingerprints used to perform attribution can be catastrophically affected by perturbations introduced during common content-sharing stages (e.g., uploading to content sharing platforms), such as JPEG compression and resizing. Finally, these works typically rely on costly optimization of specialized deep networks, which need thousands of samples per class to obtain acceptable results. 
    
    In this work, we adopt a forensic approach to address the above limitations of state-of-the-art works by considering the fragility of fingerprint-based approaches and the urge to address the problem in a realistic Open-Set scenario. More specifically, we take a different angle to the problem shifting from low-level, frequency-based representations to higher-level image features. Motivated by the generality and expressiveness of the representations of modern foundation multimodal models (e.g., DINOv2~\cite{oquab_dinov2_2024} and CLIP~\cite{radford_learning_2021}) that have been trained on extremely large dataset (i.e., LVD-142M~\cite{oquab_dinov2_2024}, LAION-2B\cite{schuhmann2022laion}, WIT400M~\cite{radford_learning_2021}, Datacomp-1B\cite{gadre2023datacompsearchgenerationmultimodal}, and YMFCC\cite{thomee2016yfcc100m}), we show that by using an intermediate feature respresentation scheme, we can obtain a sufficiently general representation to address synthetic image attribution in open-set scenarios. Moreover, we show that not learning the representation with an inductive bias derived by a single class of models (e.g., as in~\cite{yang_progressive_2023}) allows for detection and attribution on a more diverse set of data. Specifically, detecting synthetic images coming from different families of models and datasets. Finally, by contrast to methods that rely on model parameters (which may not be accessible), we operate in a realistic open-set setting. Our contributions can be summarized as follows:
    \begin{itemize}
        \item We explore the Open-Set model attribution of diffusion-generated images and we show that our method generalizes to diffusion-generated images much better than recent Open-Set attribution techniques (i.e.,~\cite{yang_progressive_2023, yang_deepfake_2022, bui_repmix_2022}) proposed for GAN-generated images.
        \item We propose a novel framework for the Open-Set diffusion-based model attribution task inspired by Open Set Recognition~\cite{vaze_open-set_2022}, employing a recently released large-scale dataset comprising several Diffusion Models.
        \item We show that the proposed learning framework that incorporates features extracted from foundation models (i.e., CLIP~\cite{radford_learning_2021} and DINOv2~\cite{oquab_dinov2_2024}) exceeds the performance of existing methods, even in the low-data regime, and generalizes better on images obtained from a diverse set of architectures in comparison to most of the existing works.
    \end{itemize}

\section{Related work}\label{sec:rel_work}

    \paragraph{\textbf{Attribution of synthetic images}} Attribution of synthetic images consists in assigning any given synthetic image to the generative model that produced it. Certain lines of research address this task by actively embedding a signal during the generation process~\cite{yu_artificial_2021, yu_responsible_2021, kim_decentralized_2020,fernandez_stable_2023}, by performing inverse engineering~\cite{laszkiewicz_single-model_2024, wang_where_2023, ma2023exposing, Ricker_2024_AEROBLADE}, or by conducting a forensic examination on the images~\cite{marra_gans_2019, yu_attributing_2019, frank_leveraging_2020, durall_watch_2020, girish_towards_2021, yang_deepfake_2022,bui_repmix_2022, yang_progressive_2023}. The latter (i.e., the \textit{forensic approach}) typically leads to greater flexibility, as it does not require access to the generative models (which may not be accessible). Forensic techniques used for attribution of GAN-generated images rely on unique residual patterns inherited by those generative models on the image (i.e., \textit{fingerprints}), which can be better analyzed by observing the frequency spectrum~\cite{marra_gans_2019, frank_leveraging_2020, durall_watch_2020}. However, such methods do not address diffusion-based generative models, which have emerged as the leading generative paradigm recently, whilst at the same time the adopted fingerprints used to perform attribution can be catastrophically affected by perturbations introduced during common content sharing stages (e.g., uploading to content sharing platforms), such as JPEG compression and resizing. By contrast, in this work we propose a general framework that addresses both Diffusion- and GAN-based attribution and does not suffer from fragile low-level frequency artifacts.
    
    Moreover, most existing works focus on a closed-set classification setting -- i.e., attribution is performed on images generated by a fixed and known set of models trained on a specific dataset with a specific seed and loss. This setting poses a significant limitation on the attribution mechanism, since new generative methods become available, while their models might not be open-sourced/available. To address this crucial constraint, a few recent works proposed to perform attribution at an architecture-level~\cite{yang_deepfake_2022,bui_repmix_2022} or to frame it as an open-set classification task~\cite{girish_towards_2021,yang_progressive_2023}. While these works have shown promising results, they only focus on GAN-generated images. In this work we show that, despite the fact that diffusion-based generation also inherits distinctive fingerprints in the generated images~\cite{corvi2023intriguing}, such methods do not generalize to diffusion-based generation~\cite{corvi_detection_2023, ojha_towards_2023}. By contrast, our method provides a general Open-Set~\cite{yang_progressive_2023} attribution framework where, instead of relying on fragile low-level frequency artifacts, we propose to perform attribution leveraging pre-trained features extracted from pre-trained visual encoders of powerful foundation models (i.e., CLIP~\cite{radford_learning_2021}, DINOv2~\cite{oquab_dinov2_2024}), which allows for better generalization ability to unseen model categories and a higher robustness to image perturbations.
    
    \paragraph{\textbf{Foundation models for image forensics}} While the vast majority of existing works focus address the problem of attribution on GAN-generated images, very recent works extended the analysis to synthetic images generated by Diffusion Models, but only for the simpler binary detection task. To obtain a universal fake image detection system, several works~\cite{ojha_towards_2023, sha_-fake_2023, cocchi2023unveiling, cozzolino2023raising, zhu_gendet_2023, amoroso2024parents} employed features extracted from large pretrained vision or vision-language models such as DINOv2~\cite{oquab_dinov2_2024} and CLIP~\cite{radford_learning_2021}, either in a multimodal setting by leveraging a textual prompt in conjunction with the visual information~\cite{cozzolino2023raising,sha_-fake_2023, amoroso2024parents}, or in a visual-only setting~\cite{ojha_towards_2023,cocchi2023unveiling}. The feature space is exploited by using $k$ Nearest Neighbors or linear probing approaches similarly to Ojha~\etal~\cite{ojha_towards_2023} and successive works~\cite{cocchi2023unveiling, amoroso2024parents} or by training a deep network, similarly to~\cite{zhu_gendet_2023}. However, few works~\cite{ojha_towards_2023, cozzolino2023raising, zhu_gendet_2023} evaluated their performance in the presence of unknown generators by framing the task as an outlier detection problem, and only Sha~\etal~\cite{sha_-fake_2023} evaluated the possibility of attributing Diffusion-generated images to the model that originated them, but in a limited closed-set comprising four generators. By contrast, in this work, we study the more general Open-Set Attribution setting, leveraging pre-trained features extracted from large vision encoders and focusing on generalization and robustness properties.

\section{Proposed Method}\label{sec:proposed_method}
    
\begin{figure}[t]
        \centering
        \includegraphics[width=0.99\textwidth]{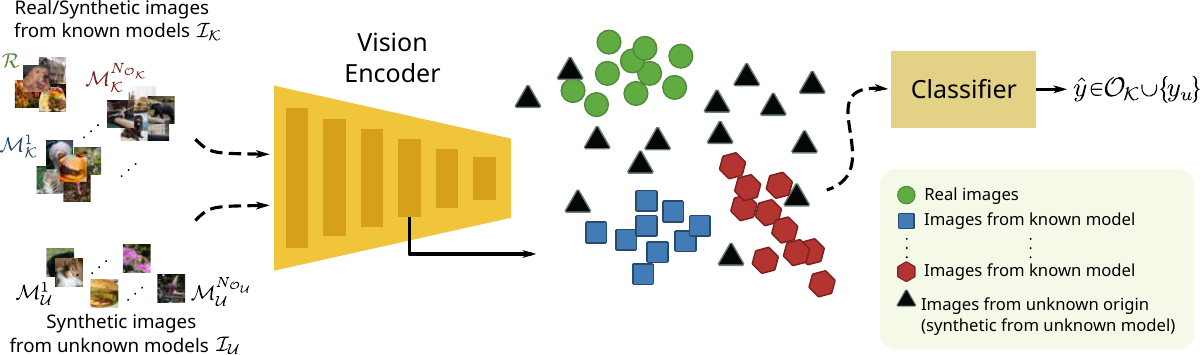}
        \caption{Overview of the proposed attribution framework -- we incorporate intermediate features from the Vision Encoder of a powerful foundation model in order to perform synthetic image detection and attribution. Using real images (from a set $\mathcal{R}$), synthetic images generated by a set of known generative models $\mathcal{O}_\mathcal{K}=\{\mathcal{M}_\mathcal{K}^1,\ldots,\mathcal{M}_\mathcal{K}^{N_{\mathcal{O}_\mathcal{K}}}\}$, and synthetic images generated by a set of unknown generative models $\mathcal{O}_\mathcal{U}=\{\mathcal{M}_\mathcal{U}^1,\ldots,\mathcal{M}_\mathcal{U}^{N_{\mathcal{O}_\mathcal{U}}}\}$, we optimize a classifier to assign images to either a known model from $\mathcal{O}_\mathcal{K}$ or ``reject'' such assignment, classifying the images as \textit{synthetic-and-unknown} ($y_u$).}.
        \label{fig:overview}
    \end{figure}

    In this section, we present our Open-Set model attribution framework. We begin by introducing the problem in \cref{ssec:problem_statement} and we present our method in \cref{ssec:method}, where we discuss the proposed feature extraction protocol and the adopted learning approaches. An overview of the proposed method is given in Fig.~\ref{fig:overview}.
    
    \subsection{Problem Statement}\label{ssec:problem_statement}
        In this work, we address the problem of synthetic image detection and attribution in the most general setting possible. More specifically, we assume that a certain set $\mathcal{I}_{\mathcal{K}}=\{x_\mathcal{K}^1,\ldots,x_\mathcal{K}^{N_\mathcal{K}}\}$ of $N_\mathcal{K}$ synthetic and real images are available and that, for each of these images model provenance (in the case of synthetic images) is known. That is, each image $x_i \in \mathcal{I}_{\mathcal{K}}$ is annotated with a label $y_i\in\{\mathcal{R}\}\cup\mathcal{O}_\mathcal{K}$, where $\mathcal{R}$ denotes the set of real images, and $\mathcal{O}_\mathcal{K}=\left\{\mathcal{M}_\mathcal{K}^1,\ldots,\mathcal{M}_\mathcal{K}^{N_{\mathcal{O}_\mathcal{K}}}\right\}$ denotes the set of $N_{\mathcal{O}_\mathcal{K}}$ known models. Assuming that at any given time a set of $N_{\mathcal{O}_\mathcal{U}}$ unkown models $\mathcal{O}_\mathcal{U}=\left\{\mathcal{M}_\mathcal{U}^1,\ldots,\mathcal{M}_\mathcal{U}^{N_{\mathcal{O}_\mathcal{U}}}\right\}$ can be used to generate new images, we take into account another disjoint set of $N_\mathcal{U}$ images $\mathcal{I}_{\mathcal{U}}=\{x_\mathcal{U}^1,\ldots,x_\mathcal{U}^{N_\mathcal{U}}\}$ generated by those unknown models $\mathcal{O}_\mathcal{U}$ -- that is, each image $x_i\in\mathcal{I}_{\mathcal{U}}$ is annotated with a label $y_i\in\mathcal{O}_\mathcal{U}$ corresponding to the unknown model that produced it.
        
        Given an image $x$ of unknown provenance, that can be either a real image or a synthetic image generated by a known or unknown model from $\mathcal{O}_{\mathcal{K}}\cup\mathcal{O}_{\mathcal{U}}$, we need to decide on whether this image is real or it has been generated by a generative model. In the latter case, we also need to predict which of the known models $\mathcal{O}_\mathcal{K}$ has been used or to ``reject'' such assignment, classifying the image as \textit{synthetic-and-unknown}, following the common practice in literature~\cite{yang_progressive_2023, chen2021adversarial, vaze_open-set_2022}. Formally, for each given image $x$, we predict its class $\hat{y}\in\mathcal{O}_\mathcal{K}\cup\{y_u\}$, where $y_u$ is a special class that indicates that the image at hand is neither real nor generated by a known model. We illustrate this process in Fig.~\ref{fig:overview}.

    \subsection{Open-Set Model Detection and Attribution}\label{ssec:method}

        \subsubsection{Visual backbone and feature extraction}\label{ssec:visual-backbones}
            In order to extract meaningful features for the task of detection and attribution, as discussed in previous sections, motivated by the generality and expressiveness of the representations of modern foundation multimodal models, we propose to employ the Vision Transformer-based~\cite{dosovitskiy2020image} encoder of a foundation model and extract intermediate features, instead of using the final feature representations. More specifically, each image $x\in\mathbf{R}^{C\times H\times W}$ is split into a sequence of square patches $\{x_i^p\}_{i=1}^N$, where $C$, $W$, and $W$ denote the channel, width, and height, respectively, and each patch has size $P\times P$. Each patch is then projected into a $d$-dimensional embedding space, and an additional token [CLS] is concatenated to the input sequence. The embedding sequence is then fed to a Transformer backbone, composed by $L$ transformer blocks, each receiving the output of the previous block. We further detail this process in \cref{ssec:experimental-setup}.

            Next, we perform the classification task (of assigning each image to a class in $\mathcal{O}_\mathcal{K}\cup\{y_u\}$ -- see Fig.~\ref{fig:overview}) following either a \textit{Linear Probe} or a \textit{$k$NN} approach as described below.
        
        \subsubsection{Linear Probe}
            In this approach, we use the extracted features to train a logistic regressor similarly to~\cite{radford_learning_2021}. By doing so, we identify the block that provides the most informative features, enabling the linear model to classify known and reject unknown models. The logistic regressor is trained with an $\ell_2$ loss using an off-the-shelf LBFGS~\cite{liu1989limited} solver. We perform a hyperparameter sweep to determine the optimal value for the regularization parameter for the features of each block. Rejection is performed based on the confidence on the prediction on each sample. This is illustrated in Fig.~\ref{fig:learning_methods} (a).
        
        \subsubsection{$\mathbf{k}$-Nearest-Neighbors ($\mathbf{k}$NN)}
\begin{figure}[t]
            \centering
            \includegraphics[width=0.99\textwidth]{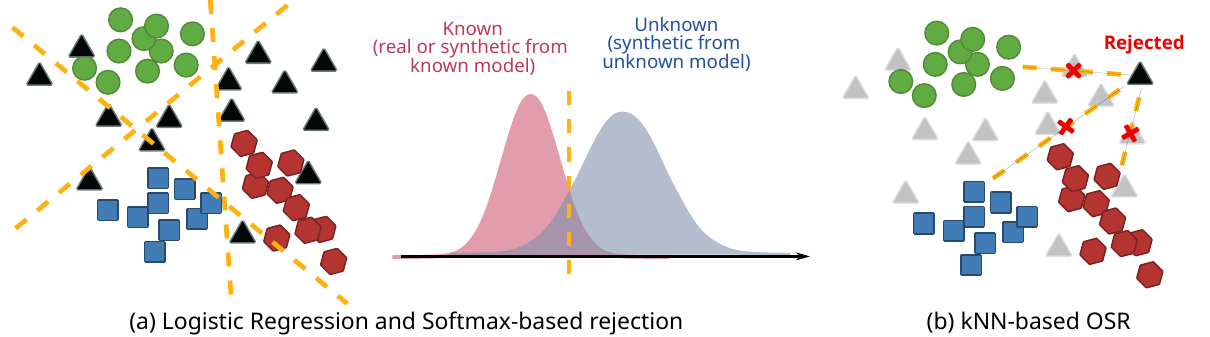}
            \caption{Proposed learning approached for Open-Set model detection and attribution.}
            \label{fig:learning_methods}
        \end{figure}

            In this approach, we measure the distances within the visual feature space, extracted by the pre-trained backbones with no further fine-tuning. As the features needed for our task may be significantly different to the original task where the backbone network has been trained for, we additionally evaluate a trained linear projection of the features using the SupCon~\cite{khosla2020supervised} loss for 10 epochs. During validation and testing, the cosine distances between each element and the training features are calculated and each data point is assigned to a specific class by majority voting on the $k$ nearest features in the embedding space. Rejection is performed based on the distance of the nearest neighbor. This is illustrated in Fig.~\ref{fig:learning_methods} (b).
        
\section{Experiments}\label{sec:experiments}

    In this section we will present the experimental evaluation of the proposed framework. In \cref{ssec:experimental-setup}, we will introduce the experimental setup. In \cref{ssec:exp-sota} present comparisons with several state-of-the-art works, and in \cref{ssec:ablation-studies} we will present ablation studies on various design choices of the proposed framework.
    
    \subsection{Experimental setup}\label{ssec:experimental-setup}
    
        \paragraph{\textbf{Dataset}} We evaluated the proposed method on the GenImage~\cite{zhu_genimage_2023} dataset, originally designed for binary Synthetic Image Detection. This dataset includes real images from ImageNet and synthetic images from eight different conditional generative models, including seven diffusion- and one GAN-based models --namely, Midjourney (MJ)~\cite{midjourney2022}, Stable Diffusion V1.5 (SD1.5)~\cite{autoamtic1111-stable-diffusion-webui}, Stable Diffusion V1.4 (SD1.4) ~\cite{autoamtic1111-stable-diffusion-webui}, Wukong~\cite{wukong_2022}, VQDM~\cite{gu2022vector}, ADM~\cite{dhariwal2021diffusion}, GLIDE~\cite{nichol2021glide}, and BigGAN~\cite{brock2018large}. These generative models are either Text-to-Image (MJ, SD, Wukong, VQDM, GLIDE), utilizing ImageNet classes as text prompts, or class-conditional (ADM, BigGAN). The GenImage dataset is split into the following three subsets:
        \begin{itemize}
            \item{\textit{Seen Real}}: 4K samples of real ImageNet images,
            \item{\textit{Seen Fake}}: 16K images produced by generative models included in the training set, and
            \item{\textit{Unseen Fake}}: 16K images produced by generative models not included in the training set.
        \end{itemize}
        Following the common practice in Open Set Recognition literature~\cite{vaze_open-set_2022,yang_progressive_2023}, we built five different splits of the dataset by varying the models included in \textit{Seen Fake} and \textit{Unseen Fake} subsets. An example of such split is shown in \cref{tab:genimage_split}. We provide the details for all five splits in \cref{sec:supmat-genimage}. In our main experiments we used $n=4$ models from both \textit{Seen Fake} and \textit{Unseen Fake}. An ablation on the number of $n$ of training classes (models) can be found in \cref{ssec:ablation-studies}. We train all competing methods~\cite{yang_progressive_2023, bui_repmix_2022, yang_deepfake_2022} on the five dataset splits and report the average of each metric in the results.
        
        \begin{table}[tb]
            \caption{Example split of the GenImage~\cite{zhu_genimage_2023} dataset similarly to~\cite{vaze_open-set_2022,yang_progressive_2023}.}
            \centering
            \setlength{\tabcolsep}{10pt} 
            \renewcommand{\arraystretch}{1.05} 
            \begin{tabular}{l|c|c}
            \hline
            Image Source & Set &{Generator}\\
            \hline
            ImageNet~\cite{deng2009imagenet} & Seen Real & -\\
            \hline
            \multirow{8}{*}{GenImage~\cite{zhu_genimage_2023}} & \multirow{4}{*}{Seen Fake} &wukong~\cite{wukong_2022}\\
            & & Midjourney~\cite{midjourney2022} \\
            & & SD1.4~\cite{autoamtic1111-stable-diffusion-webui}\\
            & & VQDM~\cite{gu2022vector}\\
            \cline{2-3}
            & \multirow{4}{*}{Unseen Fake} &glide~\cite{nichol2021glide}\\
            & & ADM~\cite{dhariwal2021diffusion} \\
            & & SD1.5~\cite{autoamtic1111-stable-diffusion-webui}\\
            & & BigGAN~\cite{brock2018large}\\
            \hline
            \end{tabular}
            \label{tab:genimage_split}
        \end{table}
        
        \paragraph{\textbf{Visual backbones and feature extraction}}
            We evaluated two families of backbones pre-trained on large-scale datasets for feature extraction, both adopting the Vision Transformer~\cite{dosovitskiy2020image} architecture. The first was trained on a cross-modal setting using the contrastive loss, while the second was trained on a visual-only setting using self-supervision. Specifically, we employed the vision encoder of CLIP~\cite{radford_learning_2021}, using the OpenCLIP\footnote{https://github.com/mlfoundations/open\_clip} implementation trained on the LAION-2B dataset, composed of 2 billion image-text couples collected from the Internet,
            and DINOv2~\cite{oquab_dinov2_2024}, which has been pre-trained on the LVD-142M dataset using a self-supervision approach based on image augmentation and enforcing consistency between local and global patches of the image. More specifically, we employed the base version of ViT~\cite{dosovitskiy2020image} for all considered backbones, which is composed by $L=12$ Transformer blocks and $768$-dimensional embeddings. The pre-trained visual backbones were used as feature extractors for each image during the training, validation, and testing phases. Each image $x \in \mathbf{R}^{C \times H \times W}$ was split into a sequence of squared patches $\{x_i^p\}_{i=1}^N$, where $C=3$, $W=224$ and $W=224$ are the channel, with, and height, respectively, and each patch has size $16\times 16$ for CLIP and $14\times 14$ for DINOv2. Each patch was then projected into a $768$-dimensional embedding and an additional token (i.e., \textit{[CLS]}) was concatenated to the input sequence. The embedding sequence was then fed to a Transformer backbone, composed by $L=12$ transformer blocks, each receiving the output of the previous block.
        
        \paragraph{\textbf{Evaluation metrics}} Following the common practice in the relevant literature, we evaluated the proposed framework using the standard metrics proposed for Open Set Recognition~\cite{vaze_open-set_2022} and Open Set Model Attribution~\cite{yang_progressive_2023}. More specifically, we report results on the accuracy on the Closed-Set setting, composed of \textit{Seen Real} and \textit{Seen Fake} images, and on the Area Under the ROC curve (AUROC) on the Open-Set setting, composed of \textit{Seen Real}, \textit{Seen Fake}, and \textit{Unseen Fake} images. The AUROC is a threshold-independent metric that monitors the true positive rate against the false positive rate by varying the rejection threshold, and it can be interpreted as the probability that a positive example is assigned a higher detection score than a negative example.
        
        To balance between the Closed-Set accuracy and the Open-Set rejection, we further employed the Open-Set Classification Rate (OSCR)~\cite{dhamija_reducing_2018} metric. The OSCR metric is defined using the Correct Classification Rate (CCR), defined as the fraction of examples of \textit{Seen data} $\Ds$ where the correct class $\hat{k}$ has the maximum probability and the probability is greater than a threshold $\tau$, 
        \begin{equation}
            \mathrm{CCR}(\tau) = \frac{\bigl|\{x \mid x \in \Ds \wedge \argmax_k P(k\,|\,x) = \hat k \wedge P(\hat{k} | x) > \tau \}\bigr|}{|\Ds|},
        \end{equation}
        and the False Positive Rate (FPR), defined as the fraction of samples from unknown data $\Du$ that are classified as any known class $k$ with a probability greater than a threshold $\tau$,
        \begin{equation}
            \mathrm{FPR}(\tau) = \frac{\bigl|\{x \mid x \in \Du \wedge \max_k P(k\,|\,x) \geq \tau \}\bigr|}{|\Du|}.
        \end{equation}
        The OSCR is then calculated as the area under the curve of the CCR against the FPR metrics. Following the protocol proposed in~\cite{neal2018open}, we report the average over the five splits of the dataset.

    \subsection{Comparison with state-of-the-art (SOTA) attribution methods}\label{ssec:exp-sota}

        \subsubsection{Comparison with SOTA on diffusion-generated images}\label{ssec:exp-sota-diffusion}
            We compare our method with two common baseline architectures, fine-tuned on the GenImage~\cite{zhu_genimage_2023} dataset, and existing SOTA methods for GAN Open-Set model attribution, namely, DNA-Det~\cite{yang_deepfake_2022}, RepMix~\cite{bui_repmix_2022}, and POSE~\cite{yang_progressive_2023}. 
            
            All methods were trained on the GenImage~\cite{zhu_genimage_2023} dataset utilizing the publicly available code and the augmentations and data transformations proposed by each method.
            \begin{figure}[t]
                \centering
                \includegraphics[width=0.5\textwidth,trim={0, 0.3em, -0.5em, 0}]{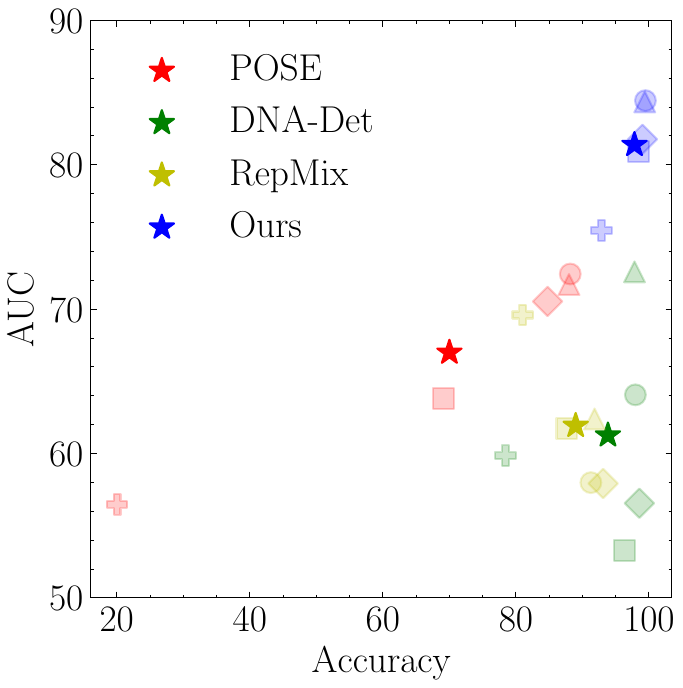}
                \begin{minipage}[b]{0.45\textwidth}
                \caption{\textbf{Comparison with GAN Attribution methods on closed-set performance (Accuracy) and open-set performance (AUROC)} on the GenImage~\cite{zhu_genimage_2023} dataset. Foreground points in bold show results averaged across five ``Seen/Unseen'' class splits for each method (following standard practice in the OSR literature), while background points, shown faint, indicate results from the underlying individual splits. Our method outperforms state-of-the-art in both closed-set and open-set performance.}
                \label{fig:genimage-attribution-comparison}
              \end{minipage}
            \end{figure}
            \cref{fig:genimage-attribution-comparison} highlights that our method demonstrates superior attribution performance both in Closed-Set attribution, assessed by Accuracy, and in the Open-Set, assessed by AUROC.
            \begin{table}[t]
            \caption{Diffusion Models Open Set Origin Attribution on GenImage~\cite{zhu_genimage_2023} dataset. We report the Closed-Set Accuracy and Open-Set AUC and OSCR. We report the mean on the five splits of the dataset $\pm$ the standard deviation. The best performance is highlighted in \textbf{bold} and second best is \underline{underlined}.}
              \setlength{\tabcolsep}{3.5pt}
              \renewcommand{\arraystretch}{1.05} 
              \centering
              \begin{tabular}{lccc}
              Method  & Acc. & AUC & OSCR\\
              \midrule
              ResNet-50~\cite{he2016deep} & $91.66 \pm 3.74$  & $70.74 \pm 5.27$ & $68.29 \pm 6.98$\\
              ViT~\cite{dosovitskiy2020image} & $93.26 \pm 2.83$ & $67.77 \pm 5.58$ & $66.02 \pm 7.24$ \\
              \midrule
              DNA-Det~\cite{yang_deepfake_2022} & $93.83 \pm 7.72$ & $61.27 \pm 6.70$  & $75.08 \pm 11.02$\\
              RepMix~\cite{bui_repmix_2022} & $88.98 \pm 4.36$ & $61.93 \pm 4.26$ & $57.92 \pm 1.73$\\
              POSE~\cite{yang_progressive_2023} & $70.00 \pm 25.95$ & $67.00 \pm 6.08$ & $53.35 \pm 19.82$\\
              \midrule
              \textbf{Ours} (NN) & $93.26 \pm 5.46$ & $77.18 \pm 1.88$ & $72.06 \pm 3.26$\\
              \textbf{Ours} (NN+) & $\underline{94.59 \pm 5.92}$ & $\underline{80.96 \pm 4.85}$ & $\underline{77.80 \pm 9.27}$\\
              \textbf{Ours} (LP) & $\mathbf{97.82 \pm 2.51}$ & $\mathbf{81.39 \pm 3.28}$ & $\mathbf{80.78 \pm 4.02}$\\
              \bottomrule[1pt]
              \end{tabular}
              \label{tab:genimage-baseline-comparison}
              \vspace*{-\baselineskip}
            \end{table}
            
            In \cref{tab:genimage-baseline-comparison} we show the performance comparison between our method and the baselines and SOTA. While DNA-Det~\cite{yang_deepfake_2022} and RepMix~\cite{bui_repmix_2022} achieve closed-set accuracy close to or above 90\%, their open-set performance is inferior to ours. Conversely, POSE~\cite{yang_progressive_2023} exhibits improved open-set performance compared to other GAN attribution methods, but falls significantly short in closed-set accuracy. In addition, POSE's training mechanism, based on the interpolation of diverse fingerprints, is fragile to variations of classes seen during training, as highlighted by the high standard deviation. Fine-tuned deep architectures, such as ResNet-50~\cite{he2016deep} and ViT~\cite{dosovitskiy2020image}, show better open-set recognition performance compared to GAN-specific methods, highlighting the low effectiveness of the training techniques proposed in GAN attribution methods when transferred to Diffusion-generated images.
            
            Our approach is in general superior with linear probing (LP) getting the highest gain in term of all three metrics, with gains of $4\%$, $10\%$, and $5\%$ in terms of Accuracy, AUC, and OSCR, respectively, with respect to the best SOTA performer for each metric. Moreover, the standard deviation across the splits is lower than the best performing competitor in each setting, showing the robustness of our findings. The incorporation of the SupCon~\cite{khosla2020supervised} (NN+) yields further improvement for the simpler Nearest Neighbour approach (NN). Simpler baselines, i.e., ResNet-50 and ViT, still obtain competitive results in term of accuracy but fail to deliver performance in the Open-Set scenario, as shown by the AUC and OSCR metrics.

        \subsubsection{Comparison with SOTA on GAN-generated images}\label{ssec:exp-sota-gan}
            \begin{table*}[t!]
            \renewcommand{\arraystretch}{1.10} 
            \caption{GAN Open Set Origin Attribution on OSMA~\cite{yang_progressive_2023} dataset. Similar to \cref{tab:genimage-baseline-comparison}, we report the Closed-Set Accuracy and Open-Set AUC and OSCR for the unknown data. Results are averaged on the five splits of the dataset. The best performance is highlighted in \textbf{bold}, second best is \underline{underlined}.}
            \centering
            \begin{tabular}{lccccccccccccc}
            \toprule
            \multirow{2}{*}{Method} & \multirow{2}{*}{Acc.} & & \multicolumn{2}{c}{Unseen Seed} & & \multicolumn{2}{c}{Unseen Arch.} & & \multicolumn{2}{c}{Unseen Data} & & \multicolumn{2}{c}{Unseen All} \\  
            \cmidrule(lr){4-5} \cmidrule(lr){7-8} \cmidrule(lr){10-11}  \cmidrule(lr){13-14}  
             & & & AUC & OSCR && AUC & OSCR && AUC & OSCR && AUC & OSCR \\
            \midrule
            PRNU~\cite{marra_gans_2019} & 55.27 && \textbf{69.20} &	49.16 &&	70.02 &	49.49&& 67.68 &	48.57 && 68.94 & 49.06\\
            Yu~\etal~\cite{yu_attributing_2019} & 85.71 && 53.14 & 50.99 &&	69.04 &	64.17 && 78.79 & 72.20 && 69.90 & 64.86\\
            DCT-CNN~\cite{frank_leveraging_2020} & 86.16 &&	55.46 &	52.68 &&	72.56 &	67.43 && 72.87 & 67.57 && 69.46 & 64.70\\
            DNA-Det~\cite{yang_deepfake_2022}  & 93.56 && 61.46 &	\underline{59.34} && \underline{80.93} & 76.45 && 66.14 & 63.27 && 71.40 & 68.00\\
            RepMix~\cite{bui_repmix_2022} & 93.69 && 54.70 & 53.26 && 72.86 &	70.49 && 78.69 & 76.02 && 71.74 & 69.43 \\
            POSE~\cite{yang_progressive_2023} & \underline{94.81} &&	\underline{68.15} &	\textbf{67.25} && \textbf{84.17} &	\textbf{81.62} && 88.24 & 85.64 &&	\textbf{82.76} & \textbf{80.50}\\ 
            \midrule
            \textbf{Ours} (NN) & 94.18 && 57.62 & 56.69 && 77.95 & 75.16 && 90.60 & \textbf{91.77} && 80.42 & 77.11\\
            \textbf{Ours} (NN+) & 95.31 && 56.18 & 54.04 && 80.22 & 77.27 && \underline{88.90} & 85.42 && \underline{80.65} & 78.22\\
            \textbf{Ours} (LP) & $\mathbf{97.29}$ && 54.15 & 54.00 && 78.78 & \underline{78.12} && \textbf{90.60} & \underline{89.52} && $79.29$ & $\underline{78.77}$\\
            \bottomrule
            \end{tabular}
            \label{tab:osma-baseline-comparison}
            \vspace*{-\baselineskip}
            \end{table*}
            
            To validate the generalization ability of our method in attributing images produced by other classes of models, in \cref{tab:osma-baseline-comparison} we compare our work against GAN attribution SOTA works on the OSMA~\cite{yang_progressive_2023} dataset, which consists of images generated by several GAN architectures. Although in this setting POSE~\cite{yang_progressive_2023} achieves a better open-set performance, our method achieves a better closed-set accuracy and an open-set performance comparable with other GAN attribution methods. In this setting, the $k$NN-based approach with a trained linear projection achieves a better AUC, while the best accuracy and Open-Set OSCR is achieved by linear probe.
            We provide extended results with standard deviations in \cref{sec:osma-extended}.
    
    \subsection{Ablation studies}\label{ssec:ablation-studies}

\begin{figure}
            \centering
            \includegraphics[width=0.9\textwidth]{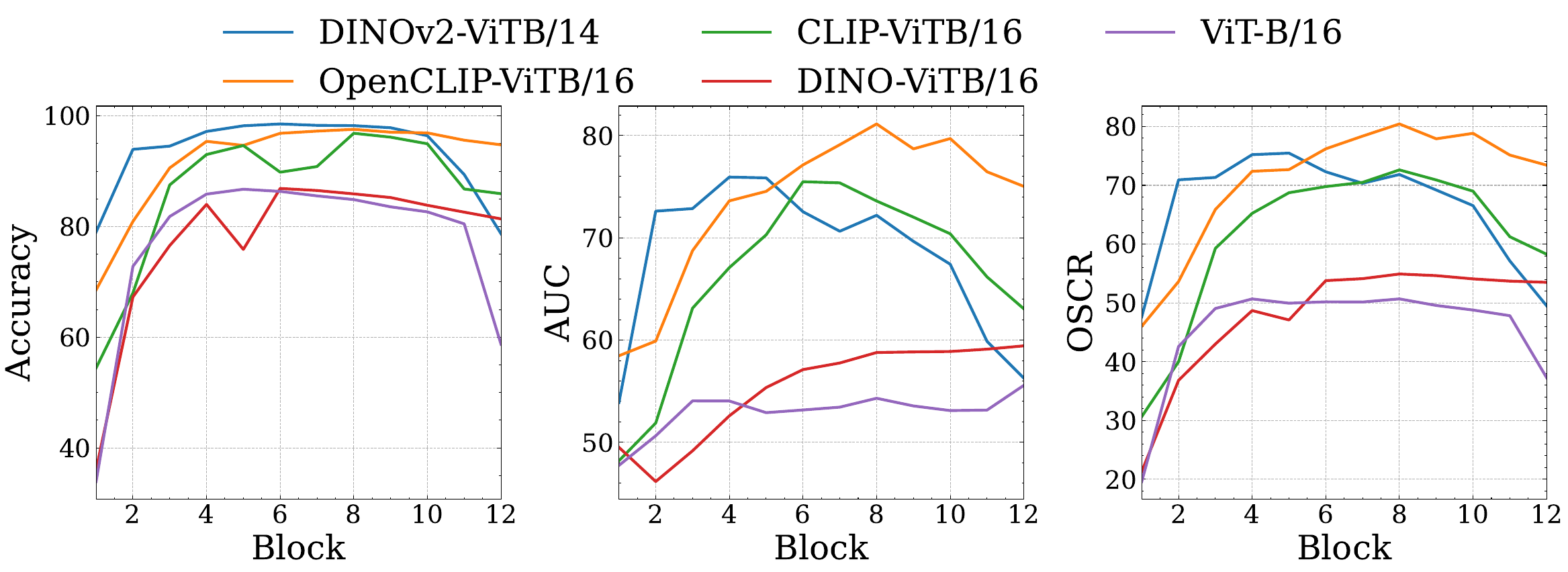}
            \caption{Comparison of closed-set and Open Set performance on GenImage of a linear probe trained with features extracted from different backbones with the $n$-th Transformer block of each backbone. Results are averaged across the five splits of the dataset.}
            \label{fig:architecture-ablation}
            \vspace*{-\baselineskip}
        \end{figure}

        \paragraph{\textbf{Architecture}} We evaluated different architectures as feature extractors for linear probing. For each architecture, we performed a sweep on all available layers, evaluating closed-set and open-set performance. We evaluated five models, all sharing the ViT-B architecture. Namely, the visual encoder of CLIP~\cite{radford_learning_2021}, both in the original and OpenCLIP implementation, DINOv2~\cite{oquab_dinov2_2024}, DINO~\cite{caron2021emerging}, and ViT~\cite{dosovitskiy2020image}. Results are shown in \cref{fig:architecture-ablation}. Although both CLIP and DINOv2 attain a closed-set accuracy above 95\%, they vary greatly in open-set performance, with OpenCLIP reaching the best AUC and OSCR. Using features from the last layer of all models lead to a lower closed-set accuracy, as well as Open-Set performance. Therefore, the choice of the point of the network to be used for feature extraction is important for the final classifier performance, as the backbone will not be fine-tuned for the attribution task. DINOv2 shows the best Open Set performance in the very first layers, and it sharply decreases in successive layers. 

\begin{figure}
            \centering
            \includegraphics[width=0.9\textwidth]{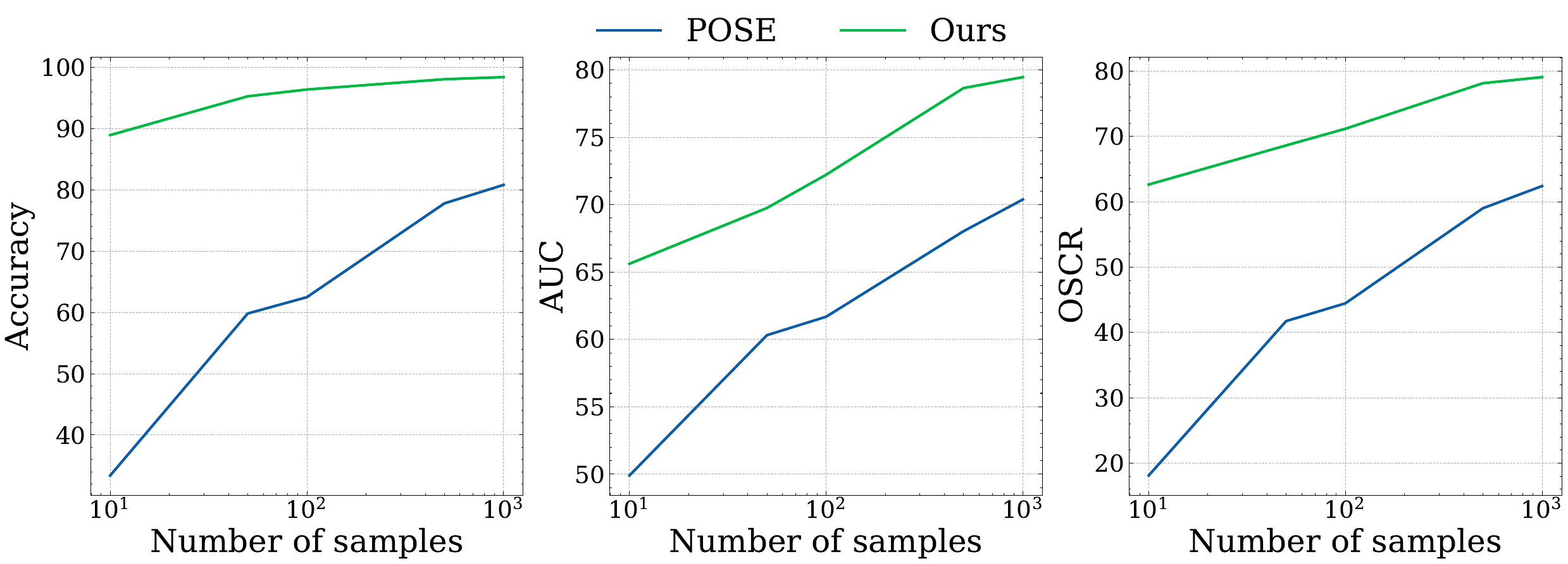}
            \caption{Closed-Set Accuracy and Open Set performance of classifiers trained on an increasing number of samples per class, ranging from 10 to 4k.}
            \label{fig:few-shot}
            \vspace*{-\baselineskip}
        \end{figure}

        \paragraph{\textbf{Impact of number of training samples}} We evaluated our approach in the scenario when only a limited number of data samples is available. Specifically, we employed a number ranging from 4K to a few-shot scenario with only 10 samples for each of the 5 known classes of GenImage~\cite{zhu_genimage_2023}. We show in \cref{fig:few-shot} that our method achieves a good OSCR performance even in the most challenging configuration, while POSE's~\cite{yang_progressive_2023} OSCR score drops to under 20\% in this setting. As expected, using pre-trained features from a large-scale model as CLIP provide great data efficiency, requiring only a limited number of data points for training.

        \paragraph{\textbf{Number of training classes}} In \cref{tab:num-classes} we compare our methods with SOTA when the number of training classes is varied between 2 and 8, with the remaining classes being inserted in the Unknown Fake set. Our method shows a consistent performance in all considered scenarios, while POSE's~\cite{yang_progressive_2023} performance drops significantly when decreasing the number of training classes to just 2. The Closed-Set setup with $8$ generators appears to be the most challenging, as it requires to distinguish models with very similar characteristics (e.g., SDv1.4 and SDv1.5). RepMix~\cite{bui_repmix_2022} also achieves a consistent performance, but shows a lower accuracy and open-set performance. Since our method does not rely on the simulation of diverse fingerprints~\cite{yang_progressive_2023}, it performs well even with few classes.
        
        \begin{table}[t]
            \centering
            \renewcommand{\arraystretch}{1.10} 
            \caption{Closed-Set Accuracy and Open Set performance of classifiers trained on GenImage~\cite{zhu_genimage_2023} when varying the ratio of known and unknown classes. The following scenarios are considered: 2 classes, 4 classes, 6 classes, and 8 classes (close-set only). The best performance is highlighted in \textbf{bold}.}
            \begin{tabular}{l@{\hskip 0.05in}cccc@{\hskip \tabcolsep}cccc@{\hskip \tabcolsep}cccc@{\hskip \tabcolsep}c}
            \toprule[1pt]
            \multirow{2}{*}{Method} & \multicolumn{3}{c}{2 Classes} && \multicolumn{3}{c}{4 Classes} && \multicolumn{3}{c}{6 Classes} && \multicolumn{1}{c}{8 Classes} \\
            \cmidrule(lr){2-4}\cmidrule(lr){6-8}\cmidrule(lr){10-12}\cmidrule(lr){14-14} & Acc. & AUC & OSCR && Acc. & AUC & OSCR && Acc. & AUC & OSCR && Acc.\\
            \midrule
            RepMix~\cite{bui_repmix_2022} & $85.24$ &$65.85$ & $63.40$ && $88.98$ &$61.93$ &$57.92$ && $\mathbf{93.48}$ &$66.31$ & $55.44$ && $76.10$\\
            POSE~\cite{yang_progressive_2023} & $65.51$ &$59.02$ &$41.75$ && $70.00$ &$67.00$ &$53.35$ && $80.93$ &$66.87$ &$61.58$ && $63.39$\\
            \midrule
            \textbf{Ours} (LP) & $\mathbf{98.40}$ &$\mathbf{82.70}$ &$\mathbf{82.33}$ && $\mathbf{97.82}$ &$\mathbf{81.39}$ &$\mathbf{80.78}$ && $87.71$ &$\mathbf{79.77}$ &$\mathbf{73.65}$ && $\mathbf{85.54}$\\
            \bottomrule[1pt]
            \end{tabular}
            \label{tab:num-classes}
            \vspace*{-\baselineskip}
        \end{table}

\begin{figure}
            \centering
            \includegraphics[width=0.7\linewidth]{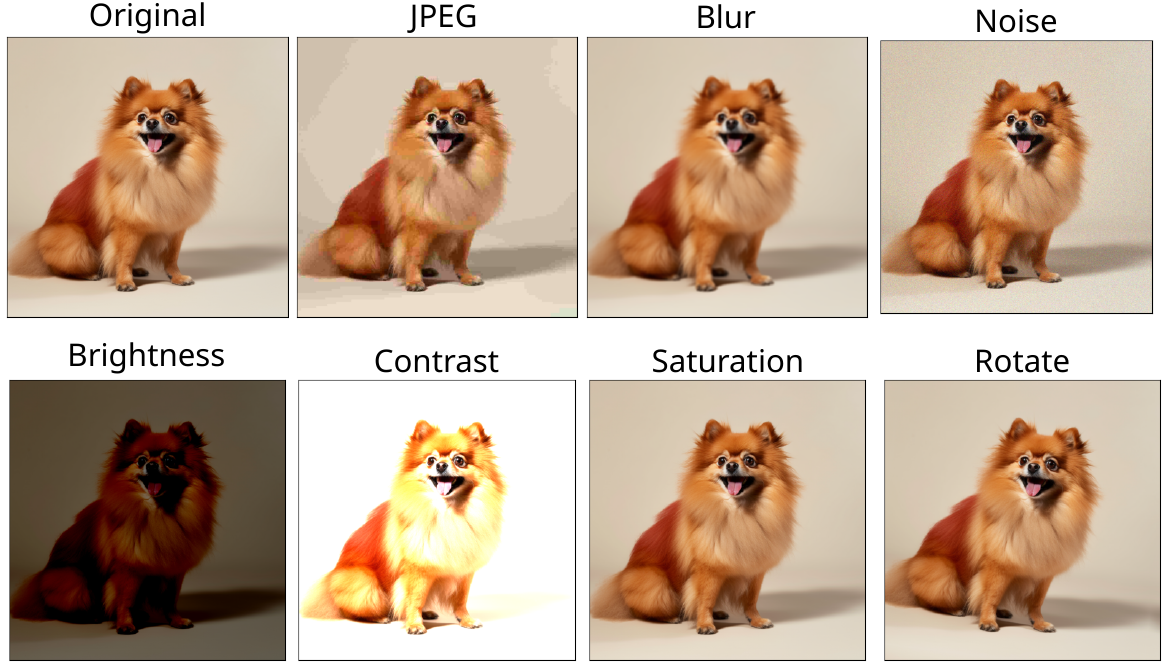}
            \caption{Visual comparison of common image perturbations applied on a sample image.}
            \label{fig:perturbations}
        \end{figure}

\begin{figure}
            \includegraphics[width=\linewidth]{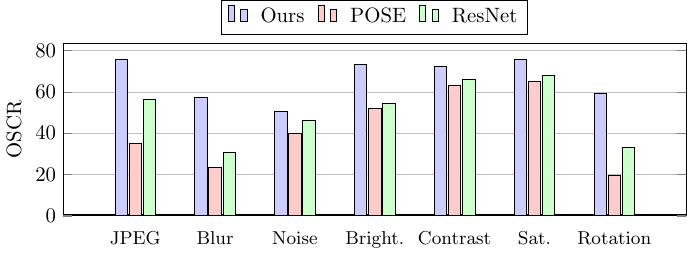}
            \caption{Robustness to perturbations on GenImage~\cite{yang_progressive_2023} dataset.}
            \label{fig:robustness}
        \end{figure}

        \paragraph{\textbf{Impact of input perturbations}} To study the effects of individual perturbations on attribution performance, we evaluated our method, POSE~\cite{yang_progressive_2023} and a baseline fine-tuned ResNet~\cite{he2016deep} on GenImage~\cite{zhu_genimage_2023} images perturbed by applying seven types of perturbations -- namely, JPEG Compression, Gaussian Blur, additive Gaussian Noise, Brightness, Contrast, Saturation, and Rotation. In \cref{fig:perturbations} we give an indicative example of the effect of each perturbation. Each perturbation is applied with an intensity randomly sampled within an interval provided in \cref{sec:supmat-perturbation}, and the methods are immunized by adding the perturbation as augmentation during the training phase. \cref{fig:robustness} reports the Closed-Set and Open-Set performance on GenImage~\cite{zhu_genimage_2023} under the attacks described above. The perturbations cause a loss in performance for all methods, with Gaussian Noise and Gaussian Blur impacting the most on our method. All perturbations that alter the structure of the image have the most impact on POSE~\cite{yang_progressive_2023}, while color alterations have a reduced impact, which is extremely reduced for our method. Remarkably, JPEG compression causes the least drop in performance in our method, while heavily affecting the existing SOTA methods. 

        \paragraph{\textbf{Rejection strategy}} While POSE~\cite{yang_progressive_2023} performs rejection based on the softmax probabilities, designing a proper score function that aims to separate ID from OOD data is essential to perform the rejection of unknown classes successfully. Thus, we further evaluated the Open-Set classification performance when using different rejection strategies, employing both post-hoc methods that rely only on the predictive distribution, and methods relying on feature statistics calculated from known training data $\mathcal{I}_{\mathcal{K}}$. We provide details of each rejection strategy in \cref{sec:supmat-rejection}. In \cref{tab:genimage-rejection-comparison} we show the Open-Set performance, measured by AUC and OSCR, of both post-hoc and methods requiring ID data on GenImage dataset. All post-hoc methods show a better performance than the MSP baseline and the best performing method is GEN~\cite{liu_gen_2023}. Methods requiring ID features, such as Mahalanobis and Residual, do not show a good performance, as they rely on ID features which are not fine-tuned for the specific task in our linear probing configuration. Overall, the best performing rejection method combines GEN and the Residual feature calculation, which improves over the MSP baseline by 4.5 points in AUROC and 2.5 points in OSCR.

        \begin{table}[bh!]
            \caption{Comparison of different rejection strategies on GenImage~\cite{yang_progressive_2023} dataset, either based on post-hoc transformation of the predictive distribution (first part) or on statistics calculated on the known data (second part).}
            \setlength{\tabcolsep}{3.5pt}
            \centering
            \begin{tabular}{lcc}
                Method & AUC & OSCR\\
                \midrule
                MSP~\cite{hendrycks_baseline_2018} & $80.19$ & $78.07$ \\
                MaxLogit~\cite{hendrycks_scaling_2022} & $80.66$ & $78.13$\\
                Energy~\cite{liu_energy-based_2020} & $80.47$ & $77.83$\\
                GradNorm~\cite{huang2021importance} & $43.92$ & $41.14$\\
                Shannon Entropy~\cite{liu_gen_2023} & $80.59$ & $78.31$\\
                GEN~\cite{liu_gen_2023} & $80.91$ & $78.41$\\ 
                GEN + Local ReAct~\cite{liu_gen_2023,wang_vim_2022} & $80.91$ & $78.40$\\
                \midrule
                Mahalanobis~\cite{lee2018simple} & $70.37$ & $65.80$\\
                Residual~\cite{wang_vim_2022} & $70.01$ & $65.40$\\
                ViM~\cite{wang_vim_2022} & $80.02$ & $77.49$\\
                GEN + ReAct~\cite{liu_gen_2023,wang_vim_2022} & $77.19$ & $74.27$\\
                GEN + Residual~\cite{liu_gen_2023} & $\mathbf{84.87}$ & $\mathbf{80.66}$\\
                \bottomrule[1pt]
            \end{tabular}
            \label{tab:genimage-rejection-comparison}
            \vspace*{-\baselineskip}
        \end{table}

\section{Conclusion}
    In this paper, we presented our method for Open-Set synthetic image detection and attribution based on foundation models. Our in-depth comparison shows that CLIP features are extremely effective in open-set attribution scenarios, while previously proposed models are designed to recognize GAN-originated signatures, which leads to reduced performance on diffusion-generated images. Moreover, CLIP performs best independently from the number of classes to be recognized in the known set and outperforms competing methods in the few-shot scenario. Finally, we analyze the robustness of our method when common image perturbations are applied. In this scenario, when immunization is provided, our method is still performing better then existing methodologies.

\vspace{0.25cm}
{\setlength{\parindent}{0cm}
\textbf{Acknowledgments:} This work was supported by the EU H2020 AI4Media No. 951911 project.
}

\clearpage
\appendix
\section{Supplementary Material}\label{supp}

The supplementary material is organized as follows:
\begin{itemize}
    \item In \cref{sec:supmat-genimage}, we provide additional details on the GenImage~\cite{zhu_genimage_2023} dataset and the splits used for attribution.
    \item In \cref{sec:supmat-frequency}, we provide a frequency analysis on the images produced by each generator.
    \item In \cref{sec:osma-extended}, we provide an extended analysis on GAN-Generated image Attribution with standard deviations.
    \item In \cref{sec:supmat-perturbation}, we provide the perturbation intensity intervals used for the robustness experiments.
    \item In \cref{sec:supmat-retrieval}, we provide retrieval results on the GenImage~\cite{zhu_genimage_2023} dataset using the $k$-NN approach described in the main paper.
    \item In \cref{sec:supmat-rejection}, provide a detailed explanation of the rejection strategies used for the linear probe classifier.
\end{itemize}

\subsection{Additional details on GenImage dataset}
\label{sec:supmat-genimage}

We provide a full description of the five splits of the GenImage~\cite{zhu_genimage_2023} dataset in \cref{tab:genimage_split1}, \cref{tab:genimage_split2}, \cref{tab:genimage_split3}, \cref{tab:genimage_split4}, and \cref{tab:genimage_split5}. 
\begin{table}[tb]
    \caption{Split 1}
    \centering
    \setlength{\abovecaptionskip}{2mm}
    \setlength{\tabcolsep}{10pt} 
    \renewcommand{\arraystretch}{1.25} 
    \begin{tabular}{l|c|c}
    \hline
    Image Source & Set &{Generator}\\
    \hline
    ImageNet~\cite{deng2009imagenet} & Seen Real & -\\
    \hline
    \multirow{8}{*}{GenImage~\cite{zhu_genimage_2023}} & \multirow{4}{*}{Seen Fake} &wukong~\cite{wukong_2022}\\
    & & Midjourney~\cite{midjourney2022} \\
    & & SD1.4~\cite{autoamtic1111-stable-diffusion-webui}\\
    & & VQDM~\cite{gu2022vector}\\
    \cline{2-3}
    & \multirow{4}{*}{Unseen Fake} &glide~\cite{nichol2021glide}\\
    & & ADM~\cite{dhariwal2021diffusion} \\
    & & SD1.5~\cite{autoamtic1111-stable-diffusion-webui}\\
    & & BigGAN~\cite{brock2018large}\\
    \hline
    \end{tabular}
    \label{tab:genimage_split1}
\end{table}

\begin{table}[tb]
    \caption{Split 2}
    \centering
    \setlength{\abovecaptionskip}{2mm}
    \setlength{\tabcolsep}{10pt} 
    \renewcommand{\arraystretch}{1.25} 
    \begin{tabular}{l|c|c}
    \hline
    Image Source & Set &{Generator}\\
    \hline
    ImageNet~\cite{deng2009imagenet} & Seen Real & -\\
    \hline
    \multirow{8}{*}{GenImage~\cite{zhu_genimage_2023}} & \multirow{4}{*}{Seen Fake} 
    & Midjourney~\cite{midjourney2022} \\
    & & SD1.4~\cite{autoamtic1111-stable-diffusion-webui}\\
    & & VQDM~\cite{gu2022vector}\\
    & & BigGAN~\cite{brock2018large}\\
    \cline{2-3}
    & \multirow{4}{*}{Unseen Fake} 
    & wukong~\cite{wukong_2022}\\
    & & glide~\cite{nichol2021glide}\\
    & & ADM~\cite{dhariwal2021diffusion} \\
    & & SD1.5~\cite{autoamtic1111-stable-diffusion-webui}\\
    \hline
    \end{tabular}
    \label{tab:genimage_split2}
\end{table}

\begin{table}[tb]
    \caption{Split 3}
    \centering
    \setlength{\abovecaptionskip}{2mm}
    \setlength{\tabcolsep}{10pt} 
    \renewcommand{\arraystretch}{1.25} 
    \begin{tabular}{l|c|c}
    \hline
    Image Source & Set &{Generator}\\
    \hline
    ImageNet~\cite{deng2009imagenet} & Seen Real & -\\
    \hline
    \multirow{8}{*}{GenImage~\cite{zhu_genimage_2023}} & \multirow{4}{*}{Seen Fake} 
    & SD1.4~\cite{autoamtic1111-stable-diffusion-webui}\\
    & & VQDM~\cite{gu2022vector}\\
    & & BigGAN~\cite{brock2018large}\\
    & & ADM~\cite{dhariwal2021diffusion} \\
    \cline{2-3}
    & \multirow{4}{*}{Unseen Fake} 
    & wukong~\cite{wukong_2022}\\
    & & glide~\cite{nichol2021glide}\\
    & & SD1.5~\cite{autoamtic1111-stable-diffusion-webui}\\
    & & Midjourney~\cite{midjourney2022} \\
    \hline
    \end{tabular}
    \label{tab:genimage_split3}
\end{table}

\begin{table}[tb]
    \caption{Split 4}
    \centering
    \setlength{\abovecaptionskip}{2mm}
    \setlength{\tabcolsep}{10pt} 
    \renewcommand{\arraystretch}{1.25} 
    \begin{tabular}{l|c|c}
    \hline
    Image Source & Set &{Generator}\\
    \hline
    ImageNet~\cite{deng2009imagenet} & Seen Real & -\\
    \hline
    \multirow{8}{*}{GenImage~\cite{zhu_genimage_2023}} & \multirow{4}{*}{Seen Fake} 
    & SD1.4~\cite{autoamtic1111-stable-diffusion-webui}\\
    & & VQDM~\cite{gu2022vector}\\
    & & BigGAN~\cite{brock2018large}\\
    & & ADM~\cite{dhariwal2021diffusion} \\
    \cline{2-3}
    & \multirow{4}{*}{Unseen Fake} 
    & wukong~\cite{wukong_2022}\\
    & & glide~\cite{nichol2021glide}\\
    & & SD1.5~\cite{autoamtic1111-stable-diffusion-webui}\\
    & & Midjourney~\cite{midjourney2022} \\
    \hline
    \end{tabular}
    \label{tab:genimage_split4}
\end{table}

\begin{table}[tb]
    \caption{Split 5}
    \centering
    \setlength{\abovecaptionskip}{2mm}
    \setlength{\tabcolsep}{10pt} 
    \renewcommand{\arraystretch}{1.25} 
    \begin{tabular}{l|c|c}
    \hline
    Image Source & Set &{Generator}\\
    \hline
    ImageNet~\cite{deng2009imagenet} & Seen Real & -\\
    \hline
    \multirow{8}{*}{GenImage~\cite{zhu_genimage_2023}} & \multirow{4}{*}{Seen Fake} 
    & SD1.5~\cite{autoamtic1111-stable-diffusion-webui}\\
    & & BigGAN~\cite{brock2018large}\\
    & & ADM~\cite{dhariwal2021diffusion} \\
    & & glide~\cite{nichol2021glide}\\
    
    \cline{2-3}
    & \multirow{4}{*}{Unseen Fake} 
    & wukong~\cite{wukong_2022}\\
    & & SD1.4~\cite{autoamtic1111-stable-diffusion-webui}\\
    & & Midjourney~\cite{midjourney2022} \\
    & & VQDM~\cite{gu2022vector}\\
    \hline
    \end{tabular}
    \label{tab:genimage_split5}
\end{table}

\subsection{Frequency analysis}\label{sec:supmat-frequency}
\begin{figure}
        \centering
        \setkeys{Gin}{width=\linewidth}
        \begin{subfigure}{0.2\textwidth}
            \caption*{\normalsize{Real}}
        \includegraphics{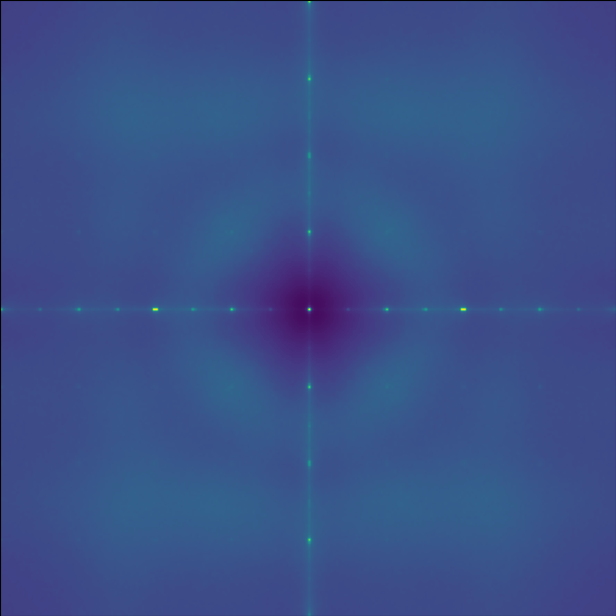}
        \end{subfigure}
        \begin{subfigure}{0.2\textwidth}
            \caption*{\normalsize{ADM}}
        \includegraphics{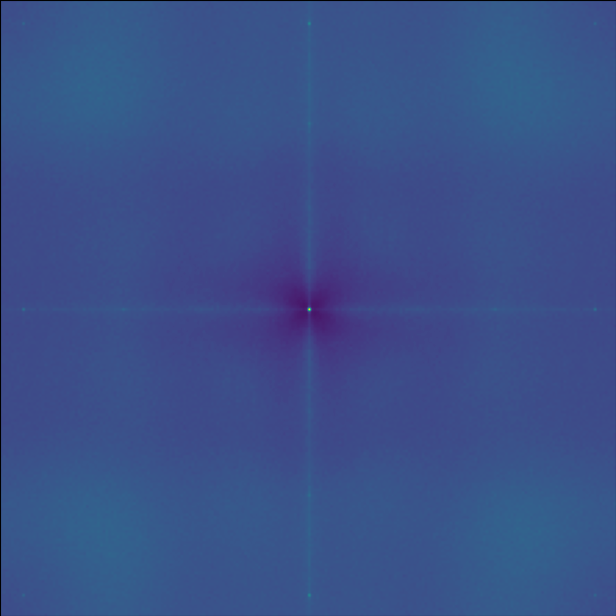}
        \end{subfigure}
        \begin{subfigure}{0.2\linewidth}
            \caption*{\normalsize{wukong}}
        \includegraphics{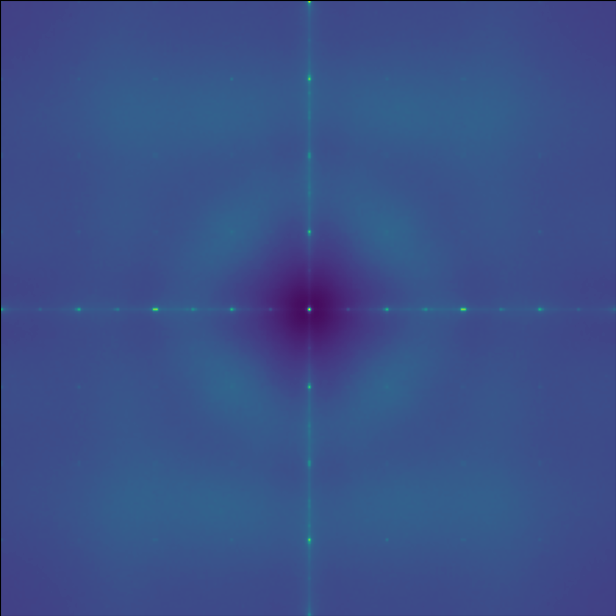}
        \end{subfigure}
        \begin{subfigure}{0.2\linewidth}
            \caption*{\normalsize{Midjourney}}
        \includegraphics{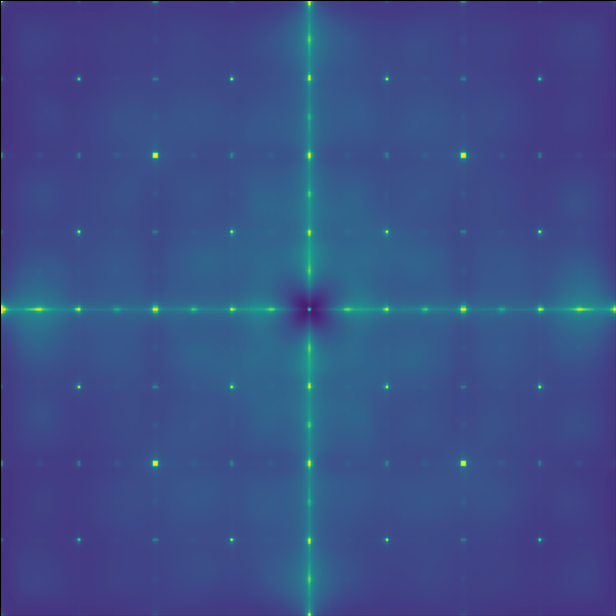}
        \end{subfigure}
        \begin{subfigure}{0.2\linewidth}
            \caption*{\normalsize{BigGAN}}
        \includegraphics{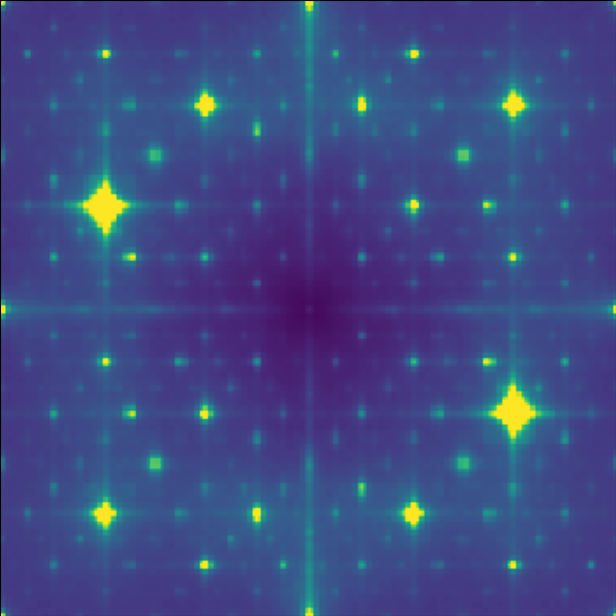}
        \end{subfigure}
        \begin{subfigure}{0.2\linewidth}
            \caption*{\normalsize{glide}}
        \includegraphics{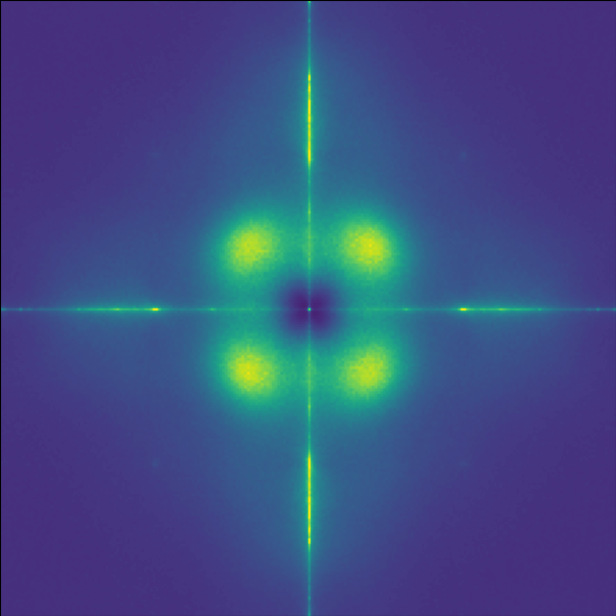}
        \end{subfigure}
        \begin{subfigure}{0.2\linewidth}
            \caption*{\normalsize{VQDM}}
        \includegraphics{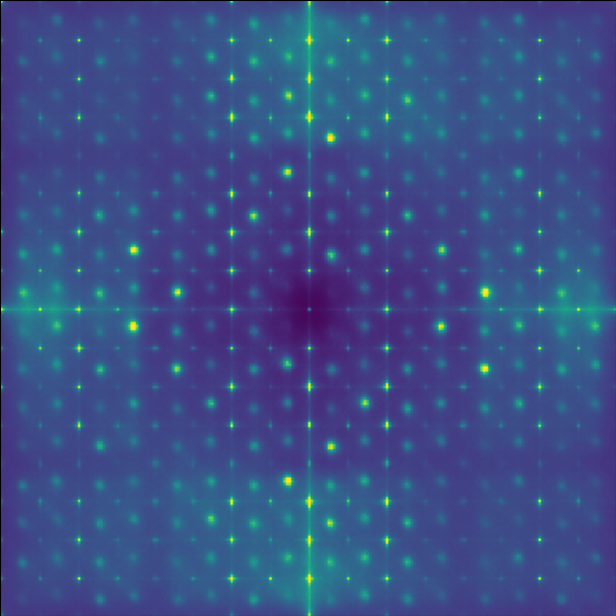}
        \end{subfigure}
        \begin{subfigure}{0.2\linewidth}
            \caption*{\normalsize{SD}}
        \includegraphics{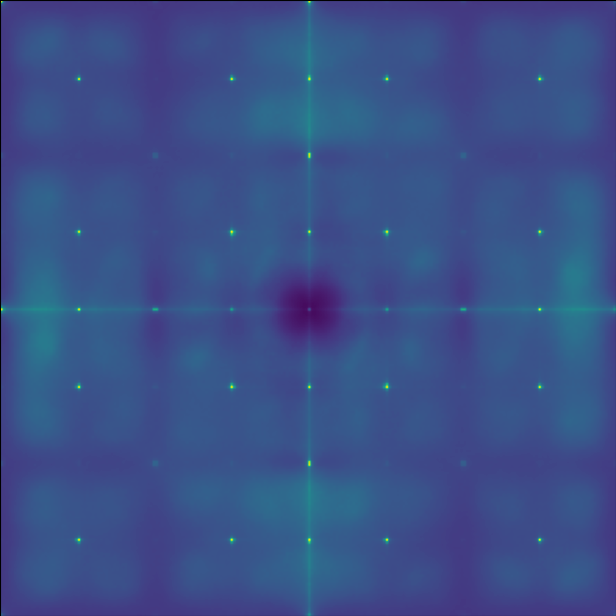}
        \end{subfigure}
        \caption{Averaged frequency spectra of synthetic images produced by different generators in GenImage dataset obtained using the pipeline described in \cite{corvi_detection_2023}}
        \label{fig:frequency-spectra}
    \end{figure}

In \cref{fig:frequency-spectra} we show a frequency analysis of the noise residuals of each class of the GenImage dataset, using the denoising model and the pipeline described in \cite{corvi2023intriguing}. While BigGAN shows the most peculiar traces, some Diffusion models (VQDM, glide, SD, Midjourney) still show visible artifacts in the frequency spectra, while others (ADM, wukong) show faint traces, almost indistinguishable from the frequency spectrum of real images.

\subsection{Extended analysis on GAN Open Set Attribution} \label{sec:osma-extended}
In \cref{tab:osma-baseline-comparison-extended}, we present an extended analysis of GAN Open Set Attribution results using the OSMA~\cite{yang_progressive_2023} dataset. The analysis includes closed-set accuracy on Seen data as well as open-set AUC and OSCR across four configurations of Unseen data. For each entry, the table reports the mean performance across five splits, along with the standard deviation. All methods achieve over 90\% accuracy on the closed-set, which consists of 15 classes, and demonstrate competitive open-set performance on three of the four Unseen data configurations (Unseen Architecture, Unseen Data, and Unseen All), with up to 64 total classes. However, the Unseen Seed configuration poses a significant challenge for all evaluated methods, and PRNU~\cite{marra_gans_2019} and POSE~\cite{yang_progressive_2023} exhibit the highest performance in this scenario. Our method, which leverages high-level semantic features, struggles in distinguishing between Seen and Unseen data under this configuration. In contrast, PRNU~\cite{marra_gans_2019} and ~\cite{yang_progressive_2023}, which utilize PRNU filters and low-level frequency information, are more responsive to the subtle traces left by model weights. To enhance performance in this challenging setting, incorporating also low-level information from the initial layers of Vision Foundation Models could be beneficial—a direction we plan to explore in future work.

\begin{table*}[t!]
    \renewcommand{\arraystretch}{1.15} 
    \centering
    \caption{GAN Open Set Origin Attribution on OSMA~\cite{yang_progressive_2023} dataset. We provide the results of three variants of our method: Nearest Neighbors (NN), Nearest Neighbors with trained projection (NN+) and Linear Probing (LP). Results are provided as the average on five splits $\pm$ the standard deviation.}
    \begin{tabular}{l@{\hskip 1em}c@{\hskip 1em}c@{\hskip 1em}c@{\hskip 1em}c}
        \toprule
        Split & Metric & \textbf{Ours} (NN) & \textbf{Ours} (NN+) & \textbf{Ours} (LP) \\
        \midrule
        Seen & Acc. & $94.18 \pm 3.68$ & $95.31 \pm 4.24$ & $\mathbf{97.29 \pm 2.35}$ \\
        \addlinespace[0.3em]
        \multirow{2}{*}{Unseen Seed} & AUC & $57.62 \pm 2.00$ & $56.18 \pm 7.42$ & $54.15 \pm 5.21$ \\
         & OSCR & $56.69 \pm 2.29$ & $54.04 \pm 7.06$ & $54.00 \pm 5.07$ \\
         \addlinespace[0.3em]
        \multirow{2}{*}{Unseen Arch.} & AUC & $77.95 \pm 4.60$ & $80.22 \pm 1.61$ & $78.78 \pm 4.60$ \\
         & OSCR & $75.16 \pm 3.74$ & $77.27 \pm 3.98$ & $\underline{78.12 \pm 5.07}$ \\
         \addlinespace[0.3em]
        \multirow{2}{*}{Unseen Data} & AUC & $90.60 \pm 6.80$ & $\underline{88.90 \pm 6.40}$ & $\mathbf{90.60 \pm 4.39}$ \\
         & OSCR & $\mathbf{91.77 \pm 5.21}$ & $85.42 \pm 8.60$ & $\underline{89.52 \pm 5.08}$ \\
         \addlinespace[0.3em]
        \multirow{2}{*}{Unseen All} & AUC & $80.42 \pm 1.11$ & $\underline{80.65 \pm 3.28}$ & $79.29 \pm 3.68$ \\
         & OSCR & $77.11 \pm 3.40$ & $78.22 \pm 3.28$ & $\underline{78.77 \pm 4.24}$ \\
        \bottomrule
    \end{tabular}
    \label{tab:osma-baseline-comparison-extended}
\end{table*}

\subsection{Perturbation intervals}\label{sec:supmat-perturbation}
In this section we provide the perturbation intervals used in robustness experiments. Each perturbation is randomly applied to the image with $p=0.5$, and the perturbation intensity is randomly chosen within the interval. 

\begin{table}[]
    \centering
    \begin{tabular}{lccc}
        Perturbation & Parameter & Min & Max \\
        \midrule
        JPEG Compression & Quality & 10 & 90 \\
        Gaussian Blur & Kernel size & 3 & 15 \\
        Gaussian Noise & Variance & 10 & 100 \\
        Brightness & Factor & -0.5 & 0.5\\
        Contrast & Factor & 0.5 & 2.0 \\
        Saturation & Factor & 0.5 & 2.0\\
        Rotation & Degrees & 0 & 30\\
        \bottomrule
    \end{tabular}
    \caption{Perturbation intervals}
    \label{tab:perturbation-intervals}
\end{table}

\subsection{Retrieval results}\label{sec:supmat-retrieval}
We provide retrieval results using the clip embeddings without the projection layer in \cref{fig:retrieval}. Results show that using a properly chosen intermediate layer we are able to access low-level information of the image related to texture, color and focus, along with higher-level information such as the pose and semantic content.
\begin{figure}[h!]
    \centering
    \begin{center}
        \includegraphics[trim={0 15cm 0 0},clip]{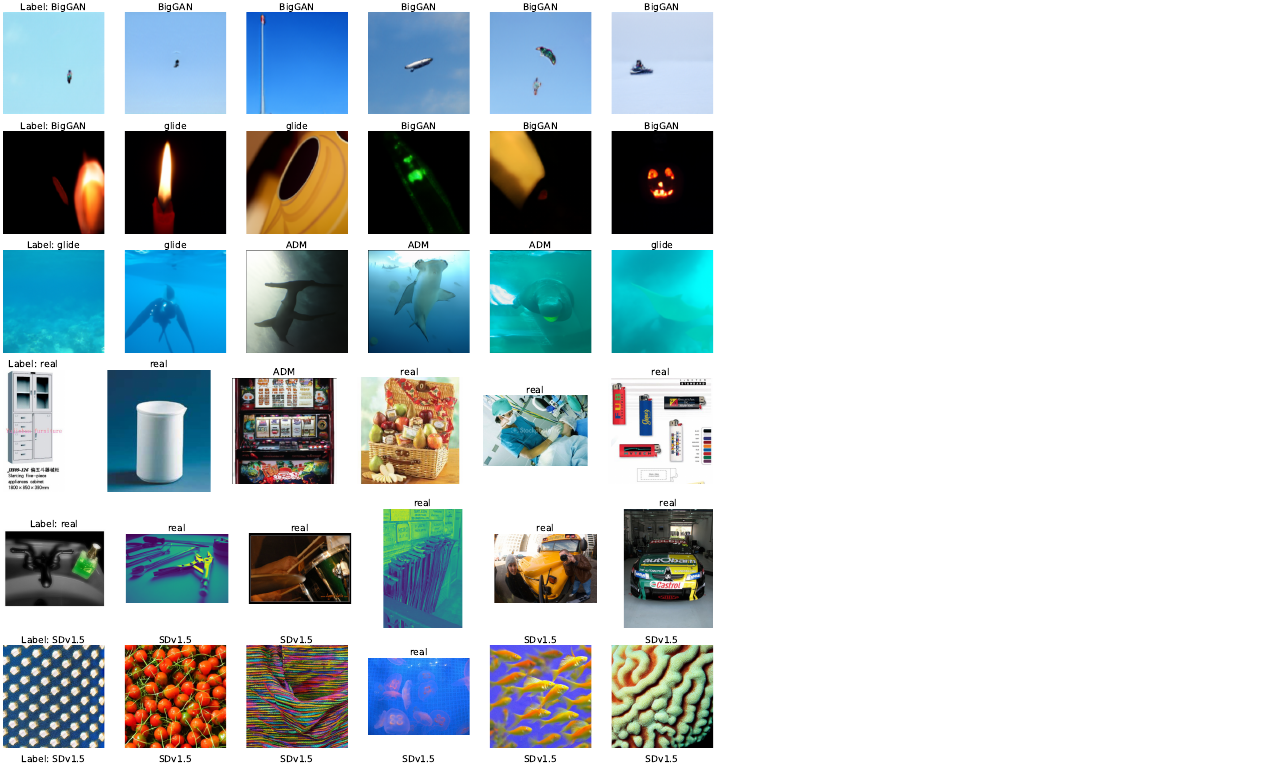}
    \end{center}
    \caption{Retrieval results on GenImage using not-projected CLIP embeddings}
\end{figure}

\begin{figure}[h!]
    \ContinuedFloat
    \begin{center}
        \includegraphics[trim={0 10cm 0 0},clip]{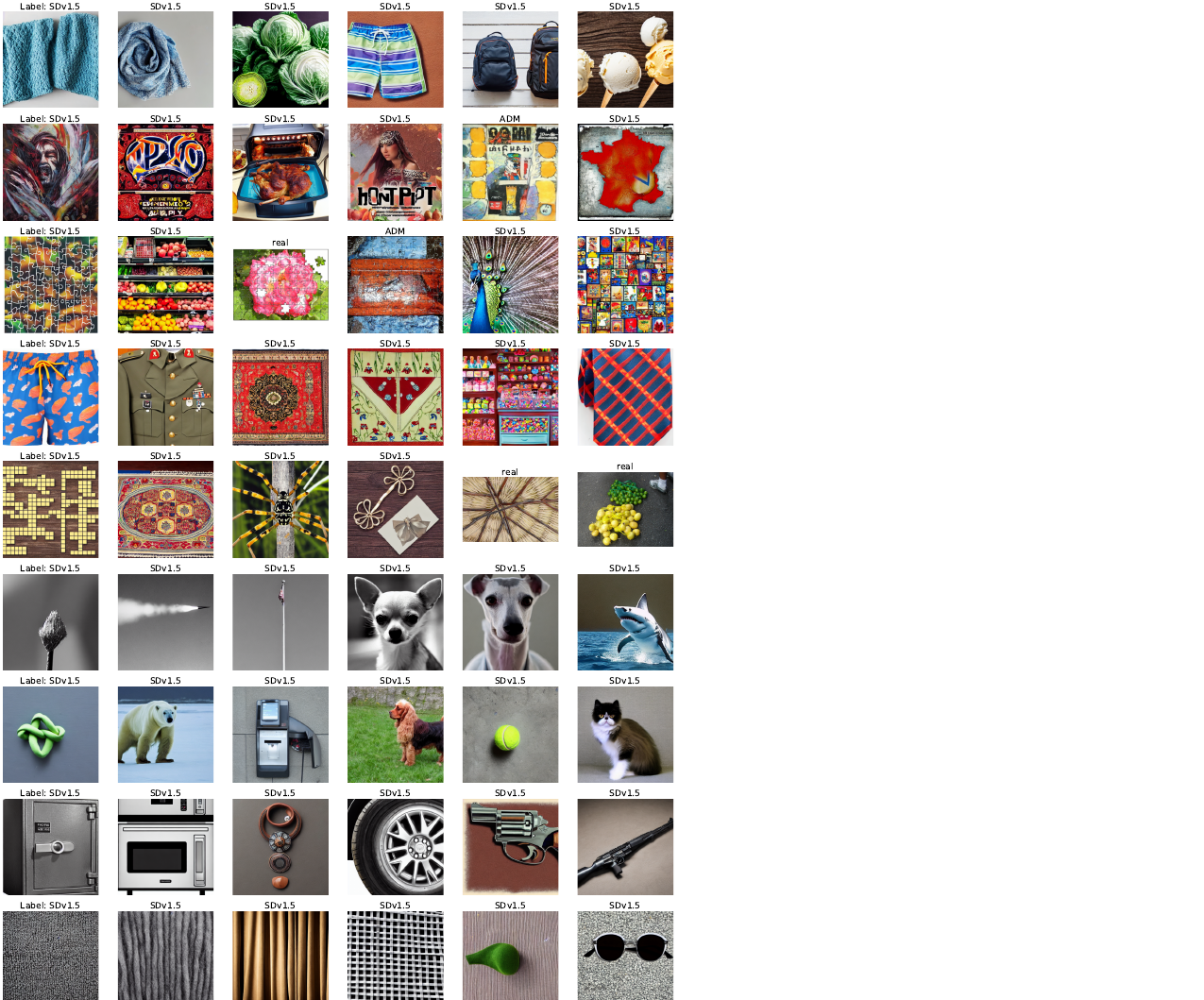}
    \end{center}
    \caption{Retrieval results on GenImage using not-projected CLIP embeddings}
    \label{fig:retrieval}
\end{figure}

\subsection{Rejection strategies}\label{sec:supmat-rejection}
One of the most simple strategies is using the maximum softmax probability (MSP)~\cite{hendrycks_baseline_2018}, which is the strategy used in POSE and in our main results. Other post-hoc methods, which require only the predictive distribution are MaxLogit~\cite{hendrycks_scaling_2022}, which uses the unnormalized logits instead of the softmax output, Energy~\cite{liu_energy-based_2020} computes the soft maximum of the logits via LogSumExp, the Shannon Entropy and the Generalized Entropy~\cite{liu_gen_2023}. We also evaluated methods that require access to the training features, such as virtual logit matching (ViM)~\cite{wang_vim_2022}, which incorporates information from the training feature space and the predictive distribution by calculating a \emph{residual} of the feature $\mathbf{z}$, $\|\mathbf{z}^{P \perp}\|$. 

\clearpage
%
\bibliographystyle{splncs04}
\bibliography{bibliography}
\end{document}